\documentclass[12pt]{article}
\usepackage{amssymb,amsmath,cite,bm}
\usepackage[dvips]{graphicx,color}
\usepackage{psfrag,subfigure}
\title{Adaptive Visual Servoing for On-Orbit Servicing}

\author{Farhad Aghili \thanks{email: farhad.aghili@concordia.ca}}

\begin{document}
%\markboth{Draft: 2018-08-24}{}
\date{}
\maketitle

\begin{abstract}
This paper presents an adaptive visual servoing framework for robotic on-orbit servicing (OOS), specifically designed for capturing tumbling satellites. The vision-guided robotic system is capable of selecting optimal control actions in the event of partial or complete vision system failure, particularly in the short term. The autonomous system accounts for physical and operational constraints, executing visual servoing tasks to minimize a cost function. A hierarchical control architecture is developed, integrating a variant of the Iterative Closest Point (ICP) algorithm for image registration, a constrained noise-adaptive Kalman filter, fault detection and recovery logic, and a constrained optimal path planner. The dynamic estimator provides real-time estimates of unknown states and uncertain parameters essential for motion prediction, while ensuring consistency through a set of inequality constraints. It also adjusts the Kalman filter parameters adaptively in response to unexpected vision errors. In the event of vision system faults, a recovery strategy is activated, guided by fault detection logic that monitors the visual feedback via the metric fit error of image registration. The estimated/predicted pose and parameters are subsequently fed into an optimal path planner, which directs the robot’s end-effector to the target’s grasping point. This process is subject to multiple constraints, including acceleration limits, smooth capture, and line-of-sight maintenance with the target. Experimental results demonstrate that the proposed visual servoing system successfully captured a free-floating object, despite complete occlusion of the vision system.
\end{abstract}

%  \cite{Aghili-2011k,Aghili-Parsa-2007b}      TRO paper
%  \cite{Aghili-Kuryllo-Okouneva-English-2010a,Aghili-Parsa-2009,Aghili-Kuryllo-Okuneva-McTavish-2009,Aghili-2008c,Aghili-Parsa-Martin-2008a}
%\cite{Aghili-Salerno-2016}
% \cite{Aghili-2010s}
%\cite{Aghili-Kuryllo-Okuneva-McTavish-2009}
%\cite{Aghili-Kuryllo-Okouneva-English-2010c,Aghili-Parsa-2008b}
%\cite{Aghili-2016c}
%\cite{Aghili-2010p}
% Aghili-2010f,Aghili-2010p,

%------------------------------------------------------
\section{Introduction}
%------------------------------------------------------

The use of autonomous robots for on-orbit servicing (OOS) has unlocked new possibilities for the commercial sector, national space agencies, and academic institutions. Servicing operations encompass a broad range of tasks, including maintenance, repairs, rescue missions, refueling, inspections, rendezvous and docking, as well as orbital debris removal~\cite{Aghili-2022,Inaba-2000,Aghili-2023,Wenberg-2020,Aghili-2011k,Wang-Meng-2020,Aghili-Parsa-2007b,Aghili-p03,Wang-Huang-2015,Aghili-2013,Aghili-Dupuis-Piedboeuf-deCarufel-1999,Aghili-Namvar-2008,Kang-Zhu-2021,Aghili-Parsa-2009b,Aghili-2020a}. A critical component of these missions is the reliable capture of target space objects using autonomous robotic arms, which must contend with non-zero relative translational and rotational motions, while adhering to multiple constraints. Many target satellites are classified as non-cooperative, as they were neither designed nor equipped for future servicing. Additionally, these objects often exhibit tumbling motion due to non-functional attitude control systems, further complicating robotic intervention. To achieve mission success, the space robot must first capture the tumbling satellite and safely mitigate its angular momentum before proceeding with repairs, rescue, or de-orbiting operations. Thus, robust visual servoing is essential for the successful servicing of unprepared satellites \cite{Mithun-2018,Aghili-2022,Liang-2022,Aghili-Parsa-2009b,Mahmood-Vagvolgyi-2020,Aghili-Kuryllo-Okouneva-English-2010a,Howard-2008}.

Despite significant advancements over the past two decades, vision-guided robotic systems continue to face challenges due to the unreliability of vision systems, environmental uncertainties, and the need to satisfy various physical and operational constraints. While image registration processes, such as the Iterative Closest Point (ICP) algorithm, have been applied in robot vision for more than 20 years, the robustness of pose tracking remains a persistent issue in computer vision applications \cite{Shang-Jasiobedzki-2005,Aghili-2022}. A key challenge is that the convergence and accuracy of image registration algorithms depend not only on the quality of the initial guess but also on the vision data, which may be compromised by factors such as sensor noise, disturbances, outliers, symmetric target views, or incomplete scan data. Therefore, the robustness and adaptability of vision-guided systems in uncertain environments are critical for autonomous robotic operations. In particular, fault tolerance is essential for designing autonomous systems that can reliably perform visual servoing tasks in space \cite{Flores-Abad-Ma-2013,NASA-2010}.

The objective of this work is to enhance the robustness and adaptability of vision-guided robotic systems for complex tasks, such as capturing free-flying objects, using a hierarchical control architecture. The proposed autonomous system can adaptively tune itself to handle not only inaccurate or erroneous visual data but also dynamic uncertainties that may affect system performance. The adaptive visual servoing system selects the most appropriate control action in response to partial or complete vision system failure. Additionally, while various vision sensors are available for robotic visual servoing, emerging 3D imaging technologies are increasingly preferred, as they do not rely on ambient lighting conditions~\cite{Samson-English-Deslauriers-Christie-2004,Aghili-Kuryllo-Okouneva-English-2010b}.

Numerous studies have explored improving the robustness of robot visual servoing for both terrestrial and aerospace applications. However, more research is needed to develop fully fault-tolerant systems that can maintain autonomous operation during partial or complete sensor failure or when vision feedback becomes inaccurate. Early work on robust robot visual servoing is reported in \cite{Zergeroglu-Dawson-2001}, which addresses planar robot manipulators with parametric uncertainty in both the mechanical dynamics and camera system. Adaptive and non-adaptive vision and force control for constrained manipulators are discussed in \cite{Baeten-DeSchutter-2002,Dean-Leon-Parra-Vega-2006,Aghili-Buehler-Hollerbach-1997a,Cheah-Hou-Zhao-2010}. Adaptive image-based visual servoing techniques have been developed to manage uncertainties in camera calibration or stereo systems \cite{Cai-Dean-2013}, and significant work has addressed system uncertainties in visual servoing control laws \cite{Xie-Low-He-2017}. Additional research has focused on managing camera field-of-view limitations for robust visual servoing \cite{Wang-Lang-Silva-2010}, while other studies have examined predictive control methods \cite{Mcfadyen-Corke-2014} and optimized trajectory planning \cite{Keshmiri-Xie-2017} for improving robustness. Path planning for industrial visual servoing tasks is discussed in \cite{Chen-Wang-Zhao-2018}. The first space-robot visual servoing system was introduced in \cite{Inaba-Oda-Hayashi-2003}, using the Japanese ETS-VII testbed. Other approaches, such as the iterative recursive least-squares and extended Kalman filter methods, have been employed to estimate the motion of floating objects from stereo or laser camera systems~\cite{Aghili-Parsa-2007b,Linchter-Dubowsky-2004,Aghili-Parsa-2009}. Motion planning and control strategies for capturing and stabilizing non-cooperative target satellites are discussed in~\cite{Luo-Sakawa-1990,Stengel-1993,Aghili-2008c,Aghili-2011k,Aghili-2016c,Aghili-2022}.

This paper presents a visual servoing system that offers fault-tolerance and self-tuning capabilities through the seamless integration of a vision registration algorithm, adaptive stochastic estimator, fault detection and recovery strategies, and optimal motion planning with predictive control \cite{Aghili-2022}. To enhance the performance and robustness of the image registration algorithm, a dynamic estimator capturing the evolution of relative motion is integrated into a closed-loop configuration with the registration process, enabling recovery after temporary vision failures. A novel dynamics model, expressed using a minimum set of inertia parameters, is developed to design a constrained estimator with equality constraints that improve system observability. The metric fit error from image registration—representing the distance between the target’s surface model and the point cloud obtained from the vision sensor—is used by a fault-detection logic for monitoring visual feedback health. The fault detection and recovery mechanisms, along with adaptive adjustments to the observation covariance matrix, allow the motion estimator to tune itself in response to inaccurate or erroneous visual data. This enables continuous pose estimation even in the presence of temporary sensor failures or brief image occlusions. The estimator provides accurate state and inertia parameter estimates to an optimal guidance and control system, facilitating precise motion planning for the robot’s end-effector, while considering multiple physical and environmental constraints.

This work introduces several significant technical contributions that improve the performance and robustness of visual servoing systems. First, the development of a novel dynamics model using a minimum set of inertia parameters enables the design of a constrained estimator that enhances system observability. Additionally, a rule-based logic system is incorporated into the motion estimation and prediction process to improve the robustness of automated rendezvous and capture. Second, the paper introduces an exact solution for optimal robot trajectories using a hard constraint method, rather than the soft constraint method based on penalty functions used in \cite{Aghili-2011k}. The penalty-function-based methods alter the original cost function of the optimal control problem by introducing an augmented performance index, while the proposed hard constraint method preserves the original cost function.  

\section*{Nomenclature}

\begin{tabular}{ll}

$\bm c_{i_k} \in \mathbb{R}^3$  &  coordinate of $i$th  point from
the point cloud \\
&  at epoch $k$ \\
$\bm d_{i_k} \in \mathbb{R}^3$  & the corresponding  point on the surface model\\
$\bm A \in \mathbb{R}^{3 \times 3}$ & rotation matrix\\
$\bm I_c \in \mathbb{R}^{3 \times 3}$ & inertia matrix\\
$\sigma_1, \sigma_2, \sigma_3$ & dimensionless inertia parameters \\
$\bm\omega$ & angular velocity \\
$\bm\mu$ & quaternion, orientation  $\{{\cal B}\}$ w.r.t $\{{\cal C}\}$ \\
$\bm\eta$ &  quaternion, orientation  $\{{\cal C}\}$ w.r.t $\{{\cal A} \}$ \\
$\bm q$ &  quaternion, orientation  $\{{\cal B} \}$ w.r.t $\{{\cal A} \}$ \\
$\bm\rho$ &  location of the grasping point expressed in $\{{\cal A} \}$ \\
$\bm\rho_o$ &  location of the CoM expressed in $\{{\cal A} \}$ \\
$\bm\varrho$ & location of the CoM expressed in $\{{\cal B} \}$ \\
$E[\cdot]$ & the expected operator\\
$\mbox{tr}(\cdot)$ & trace of a matrix\\
$\varepsilon_k$ & ICP metric fit error\\
$[\cdot \times]$ & matrix form of cross-product\\
$\otimes$ & quaternion product 
\end{tabular}

%------------------------------------------------------
\section{Fault-Tolerant Motion Estimation Based on Vision Data} \label{sec:ICP}
\begin{figure}
\psfrag{Estimator}[c][c][.8]{Estimator} \psfrag{Constrained}[c][c][.8]{Constrained}   \psfrag{Optimal}[c][c][.8]{Optimal}  \psfrag{rendezvous}[c][c][.8]{Rendezvous}
\psfrag{Trajectory}[c][c][.8]{Trajectory} \psfrag{planning}[c][c][.8]{Planning} \psfrag{Image}[c][c][.8]{Image}  \psfrag{Registration}[c][c][.8]{Registration} \psfrag{rule-base}[c][c][.8]{Rule-base} \psfrag{State Machine}[c][c][.8]{state machine}
\psfrag{eps}[l][l][.7]{$\varepsilon_k$}
\psfrag{tf}[l][l][.7]{$t_f$}
\psfrag{t0}[l][l][.7]{$t_0$}
\psfrag{r-dr-ddr}[l][l][.7]{$\bm r, \dot{\bm r}, \ddot{\bm r}$}
\psfrag{xk}[c][c][.7]{$\hat{\bm x}_k$}
\psfrag{gam}[l][l][.7]{$\gamma$}
\centering
\includegraphics[width=9.5cm]{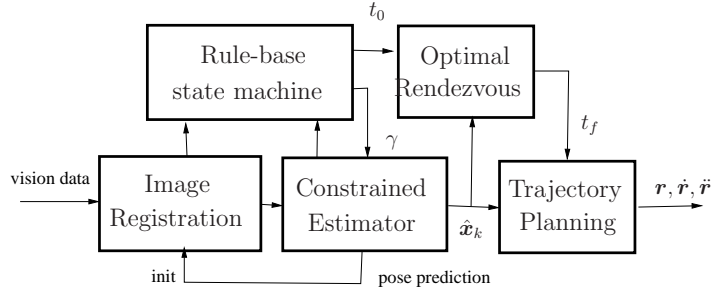} \caption{Fault-tolerant and adaptive robot vision control.} \label{fig:vision_cntr}
\end{figure}

%--------------------------------------------------------
Fig.~\ref{fig:vision_cntr} depicts  the hierarchical estimation and control approach for the fault-tolerant vision-guided robotic systems to capture moving objects. The integrated image registration algorithm and constrained noise-adaptive Kalman filter provides consistent estimation of the states and parameters of the target from the stream of visual data. The health of the visual feedback is assessed based on the metric fit error of the image registration in order to make decision whether or not the visual data has to be incorporated  in the estimation process.  Transition on the convergence of parameter estimation is monitored from the magnitude of the estimator covariance matrix while the robot remains stand-still.  The parameters are passed to the optimal guidance consisting of the optimal rendezvous and trajectory planning modules at the time of initial convergence $t_1$. The time of interception of the capturing robot and the target is determined by the optimal rendezvous module and subsequently the optimal trajectories given  physical and operational constraints are generated by the optimal planner.

The iterative closest point (ICP) algorithm is  the most popular algorithm for solving registration process of 3D CAD model and point cloud generated by a 3D vision sensor. Suppose that we are given with a set
of acquired 3D points data  ${\cal C}_k=\{\bm c_{1_k} \cdots \bm c_{m_k} \}$ at discrete time $t_k$ that corresponds to a single shape represented by model set ${\cal M}$. Here, vector $\bm c_{i_k} \in\mathbb{R}^3$  represents the coordinate of $i$th single point from the point cloud at epoch $k$. For each point $\bm c_{i_k}$, there exists at least one point on
the surface model ${\cal M}$, say $\bm d_{i_k}$, which is closer to $\bm c_{i_k}$ than other points. Therefore, one should be able to populate the data set ${\cal D}_k = \{\bm d_{1_k} \cdots \bm d_{m_k} \}$ representing  all corresponding points from an optimization process ~\cite{Simon-Herbert-Kanade-1994}. Then, the fine alignment $\{ \bar{\bm\rho}_k, \; \bar{\bm\eta}_k \}$, where $\bar{\bm\rho}_k$ represents the distance and  unit-quaternion $\bar{\bm\eta}_k$ represents the orientation, can be resolved  to minimize the distance between the data sets
 ${\cal C}_k$ and ${\cal D}_k$ through the the following least squares programming~\cite{Besl-Mckay-1992}
\begin{equation} \label{eq:min_distance}
\varepsilon_k = \mbox{arg} \min_{\bar{\bm\eta}_k, \bar{\bm\rho}_k} \frac{1}{m} \sum_{i=1}^m \| \bm A(\bar{\bm\eta}_k) \bm c_{i_k} +
\bar{\bm\rho}_k - \bm d_{i_k} \|^2.
\end{equation}
Here, variable $\varepsilon_k$ is called ICP metric fit error, and  $\bm A(\bm\eta)$ is the rotation matrix corresponding to quaternion $\bm\eta$
\begin{equation} \label{eq:R}
\bm A(\bm\eta) =\bm I + 2 \eta_o [\bm\eta_v \times] + 2 [\bm\eta_v \times]^2,
\end{equation}
where $\bm\eta_v$ and $\eta_o$ are the vector and scaler parts of the
quaternion, i.e., $\bm\eta=[\bm\eta_v^T \; \eta_o]^T$, $[\cdot \times]$ denotes the matrix form of the cross-product,
and $\bm I$ denotes  identity matrix with adequate dimension. The complete cycle of 3D registration process with incorporation of predicted pose will be later discussed in Section~\ref{sec:3D-registration}. At this point, it suffices to assume instantaneous pose at epoch $k$ as a function of the point cloud set ${\cal C}_k$, i.e.,
\begin{equation} \label{eq:ICP_outcome}
\begin{array}{l}\mbox{outcome of ICP} \\ \mbox{cycle at $ t_k $} \end{array}:= \left\{\begin{array}{c} \bar{\bm\rho}_k({\cal C}_k, \bm\rho_k^{(0)}, \bm\eta_k^{(0)}) \\ \bar{\bm\eta}_k({\cal C}_k, \bm\rho_k^{(0)}, \bm\eta_k^{(0)}) \\  \varepsilon_k \end{array} \right\}.
\end{equation}
Here,  $\bm\rho_k^{(0)}$ together with $\bm\eta_k^{(0)}$ represent initial guess for the coarse alignment to be rendered by a prediction estimation process. The remainder of this section describes development of a fault-tolerant dynamics estimator based on \eqref{eq:min_distance} and \eqref{eq:ICP_outcome}.

%=============================================================
\begin{figure}
\psfrag{rho}[l][l][.7]{$\bm\rho$}
\psfrag{rho_o}[l][l][.7]{$\bm\rho_o$}
\psfrag{vr}[l][l][.7]{$\bm\varrho$}
\psfrag{r}[l][l][.7]{$\bm r$}
\psfrag{omeg}[c][c][.7]{$\bm\omega$}
\psfrag{FA}[l][l][.7]{$\{{\cal A}\}$} \psfrag{FB}[l][l][.7]{$\{{\cal B}\}$}\psfrag{FC}[l][l][.7]{$\{{\cal C}\}$}
\centering
\includegraphics[width=9.5cm]{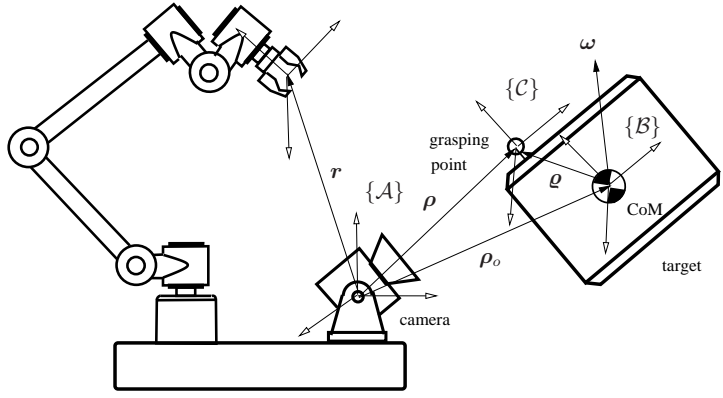} \caption{Vision-guided manipulator and target.} \label{fig:robot_target}
\end{figure}
%=============================================================

%=============================================================
%=============================================================

Fig.~\ref{fig:robot_target} illustrates definition of the coordinate frames used for the system of the vision guided manipulator and a moving target. The camera coordinate frame  is $\{{\cal A } \}$, while frames $\{{\cal B} \}$ and $\{{\cal C}\}$ are both attached to the body of the target and their origins coincide respectively with the center-of-mass (CoM) and  the
grapple fixture located at distance $\bm\varrho$ from the CoM. We also assume that frame $\{ B \}$ is aligned with principal axes of the body. The vision system gives a noisy measurement of the position and orientation of coordinate $\{{\cal C}\}$  with respect to the coordinate
frame $\{{\cal A}\}$ represented by variables $\bm\rho$ and unit quaternion $\bm\eta$. Suppose unit quaternions $\bm\mu$ and $\bm q$, respectively, represent the misalignment between coordinates $\{{\cal B}\}$ and $\{{\cal C}\}$ and the orientation
of coordinates frame $\{{\cal B}\}$ respect to $\{{\cal A}\}$.
Then, $\bm\eta$ can be considered as two
successive orientations as
\begin{equation} \label{eq:muOtimesq}
\bm\eta = \bm\mu \otimes \bm q, \quad \mbox{where} \quad  \bm\mu \otimes = \mu_o \bm I + \bm\Omega(\bm\mu_v)
\end{equation}
is the quaternion product operator, and
\begin{equation} \label{eq:Omega}
 \bm\Omega(\bm\mu_v)=\begin{bmatrix} -[\bm\mu_v
\times] & \bm\mu_v \\ - \bm\mu_v^T & 0 \end{bmatrix}.
\end{equation}
Since the target rotates with
angular velocity $\bm\omega$, quaternion $\bm\mu$ is a constant whereas $\bm\eta$ and $\bm q$ are time-varying variables. Now let us define the following non-dimensional inertia parameters
\begin{equation} \label{eq:p}
\sigma_1 = \frac{I_{yy}- I_{zz}}{I_{xx}}, \quad
\sigma_2 = \frac{I_{zz}- I_{xx}}{I_{yy}}, \quad
\sigma_3 = \frac{I_{xx}- I_{yy}}{I_{zz}},
\end{equation}
where $I_{xx}$, $I_{yy}$, and $I_{zz}$ denote the target's principal moments
of inertia. The principal moments of inertia take only  positive values and and they must also satisfy the triangle inequalities. Thus, the following inequalities are in order
\begin{align} \notag
& I_{xx}, I_{yy}, I_{zz} >0 \\ \notag
& I_{xx} + I_{yy} > I_{zz}, \quad I_{yy} + I_{zz} > I_{xx}, \quad I_{zz} + I_{xx} > I_{yy}.
\end{align}
The above inequalities imply the dimensionless inertia parameters must be bounded by
\begin{equation} \label{eq:sigma_bounds}
-1 < \sigma_1, \sigma_2, \sigma_3 < 1
\end{equation}
Also by inspection one can show that the following nonlinear constraint between the dimensionless parameters is in order
\begin{equation} \label{eq:sigma_constraint}
\Gamma(\bm\sigma) = \sigma_1 + \sigma_2 + \sigma_3 + \sigma_1\sigma_2\sigma_3  =0,
\end{equation}
which means that the dimensionless parameters are not independent. Consider the set of independent dimensionless inertia parameters $\bm\sigma = [\sigma_1 \; \;  \sigma_2]^T$. Then, from \eqref{eq:sigma_constraint} one can readily derive the third variable by
\begin{equation}
\sigma_3= - \frac{\sigma_1 + \sigma_2}{1 + \sigma_1 \sigma_2}
\end{equation}
Inequalities \eqref{eq:sigma_bounds} can be concisely described by the following vector inequality %\footnote{Vector inequality $\bm a < \bm b$ means that each component of vector $\bm a$ is less than  each corresponding component of vector $\bm b$, i.e., $a_i < b_i, \quad \forall i=1, \cdots, n.$}
\begin{equation} \label{eq:Dsigma<1}
\bm D \bm\sigma < \bm 1, \quad \mbox{where} \quad  \bm D =\begin{bmatrix} \bm I \\  -\bm I \end{bmatrix}
\end{equation}
where $\bm 1=[1\; 1\; 1 \; 1]^T$ is the vector of one . Now, we are ready to transcribe the Euler's rotation equations  in terms of the independent dimensionless parameters $\bm\sigma$, i.e.,
\begin{subequations}
\begin{equation} \label{eq:dot_omega}
\dot{\bm\omega} = \bm\phi(\bm\omega, \bm\sigma) + \bm B(\bm\sigma)
\bm\epsilon_{\tau}
\end{equation}
where $\bm\epsilon_{\tau} = \bm\tau/{\rm tr}(\bm I_c)$ is the angular acceleration disturbance, $\bm I_c =\mbox{diag}(I_{xx}, I_{yy},I_{zz})$ is the inertia tensor,  ${\rm tr}(\cdot)$ is the trace operator,
\begin{align} \notag
\bm B(\bm \sigma) & = \bm I + \begin{bmatrix}\frac{2+\sigma_1 \sigma_2 + \sigma_1}{1 - \sigma_2} & 0 & 0 \\
0 & \frac{2+\sigma_1 \sigma_2 - \sigma_2}{1 + \sigma_1} & 0 \\
0 & 0 & \frac{2+\sigma_1 -\sigma_2 }{1 + \sigma_1 \sigma_2} \end{bmatrix}, \\
\bm\phi(\bm\omega, \bm\sigma) &= \begin{bmatrix} \sigma_1 \omega_y \omega_z \\ \sigma_2 \omega_x \omega_z \\
-\frac{\sigma_1 + \sigma_2}{1 + \sigma_1 \sigma_2} \omega_x \omega_y  \end{bmatrix}.
\end{align}
\end{subequations}
Consider the following state vector pertaining to both states  and the associated dynamic parameters
\begin{equation} \label{eq:x}
\bm x=[\bm q_v^T \; \bm\omega^T \; \bm\rho_o^T \; \dot{\bm\rho}_o^T \; \underbrace{\bm\sigma^T \; \bm\varrho^T \;  \bm\mu_v^T }_{\rm parameters} \; ]^T.
\end{equation}
Then, from the well-known relationship between the time derivative of quaternion and angular rate, equations of linear and angular accelerations, and assuming constant parameters, one can describe the system dynamic in the following compact form
\begin{equation}
\dot{\bm x} = \bm f(\bm x, \bm\epsilon), \qquad \bm f(\bm x, \bm\epsilon) = \begin{bmatrix} \frac{1}{2} \mbox{vec} \big( \bm\Omega(\bm\omega) \bm q \big) \\ \bm\phi(\bm\omega, \bm\sigma) +\bm B(\bm\sigma) \bm\epsilon_{\tau} \\
\dot{\bm\rho}_o \\
\bm\epsilon_f  \\ \bm 0 \end{bmatrix},
\end{equation}
where function $\mbox{vec}(\cdot)$ returns the vector part of quaternion, $\bm\epsilon_f$ is the process noise due to force disturbance, and vector $\bm\epsilon= [\bm\epsilon_{\tau}^T \;\; \bm\epsilon_f^T]^T$ represent the overall process noise.
 Suppose $\hat{\bm q}$ represents the estimated quaternion and subsequently define small quaternion variable $\delta \bm q = \bm q \otimes \hat{\bm q}^{-1}$ to be  used as the states of linearized system, where $\hat{\bm q}^{-1}$ is quaternion inverse, i.e., $\hat{\bm q} \otimes \hat{\bm q}^{-1}=[0\; 0 \; 0 \; 1]^T$. Then, the linearized process dynamics can be described by
\begin{subequations} \label{eq:dot_deltax}
\begin{equation}
\delta \dot{\bm x} = \bm F \delta \bm x + \bm G \bm\epsilon,
\end{equation}
\begin{align} \label{Eq:F}
\bm F &=
\begin{bmatrix} - [\hat{\bm\omega} \times] & \frac{1}{2} \bm I & \bm 0 & \bm 0 & \bm 0 & \bm 0 & \bm 0 \\
\bm 0& \left(\frac{\partial\bm\phi}{\partial\bm\omega}\right)_{\!\! \hat{\bm x}}   & \bm 0 & \bm 0 & \left(\frac{\partial\bm\phi}{\partial\bm\sigma}\right)_{\!\! \hat{\bm x}}  & \bm 0 & \bm 0 \\
\bm 0 & \bm 0& \bm 0 & \bm I & \bm 0 & \bm 0 & \bm 0   \\
\bm 0 & \bm 0 & \bm 0 & \bm 0 & \bm 0 & \bm 0 & \bm 0
\end{bmatrix} \\ \label{Eq:G}
\bm G &= \begin{bmatrix} \bm 0 & \bm 0 \\
\bm B(\hat{\bm\sigma}) & \bm 0 \\
\bm 0& \bm 0\\
\bm 0& \bm I \\
\bm 0 & \bm 0
\end{bmatrix} \\
\frac{\partial\bm\phi}{\partial\bm\omega} &=
\begin{bmatrix} 0 & \sigma_1 \omega_z & \sigma_1  \omega_y \\ \sigma_2  \omega_z & 0 &
\sigma_2 \omega_x \\ -\frac{\sigma_1 +\sigma_2 }{1 + \sigma_1 \sigma_2} \omega_y & -\frac{\sigma_1 + \sigma_2}{1 + \sigma_1 \sigma_2} \omega_x & 0
\end{bmatrix} \\
\frac{\partial\bm\phi}{\partial\bm\sigma} & = \begin{bmatrix} \omega_y \omega_z & 0 \\ 0 & \omega_x \omega_z  \\ \frac{\sigma_2^2 -1}{(1+ \sigma_1 \sigma_2)^2}  \omega_x \omega_y & \frac{\sigma_1^2 -1}{(1+ \sigma_1  \sigma_2)^2} \omega_x \omega_y \end{bmatrix}.
\end{align}
\end{subequations}

Defining quaternion variations $\delta \bm\mu=\hat{\bm\mu}^{-1} \otimes \bm\mu$ and $\delta \bm\eta = \delta \bm\mu \otimes \delta \bm q$, one can readily establish the relationship between the measured quaternion and its variation through the following identity $\bm\eta = \hat{\bm\mu} \otimes \delta \bm\eta \otimes \hat{\bm q}$, and hence $\delta \bar{\bm\eta} = \hat{\bm\mu}^{-1} \otimes \bar{\bm\eta} \otimes \hat{\bm q}^{-1}$. Then, by virtue of \eqref{eq:ICP_outcome} and \eqref{eq:muOtimesq}, the observation equations can be written as
\begin{equation} \label{eq:h_nonlin}
\bm z = \begin{bmatrix} \bar{\bm\rho} \\ \delta \bar{\bm\eta}_v \end{bmatrix}= \bm h(\delta \bm x) + \bm v
\end{equation}
where vector $\bm v$ represents measurement noise, and
\begin{equation} \label{eq:h_nonlin}
\bm h (\delta \bm x) = \begin{bmatrix} \bm\rho_o + \bm A(\delta \bm q \otimes \hat{\bm q}) \bm\varrho \\ \mbox{vec} \big( \delta \bm\mu \otimes \delta \bm q \big) \end{bmatrix}.
\end{equation}
Finally, from the first-order approximation of the relationship $\bm A(\delta \bm q \otimes \hat{\bm q}) \approx 2 \bm A(\hat{\bm q})[\delta \bm q_v \times]$ and $\mbox{vec}(\delta \bm\mu \bm \otimes \delta \bm q) \approx - \delta \bm\mu_v \times \delta \bm q_v + \delta \bm q_v + \delta \bm\mu_v$, one can derive the observation sensitivity
matrix $\bm H= \bm H (\delta \hat{\bm x})$ in the following form
\begin{equation} \notag
\bm H =
\begin{bmatrix} -2\bm A(\hat{\bm q})[\hat{\bm \varrho} \times] &\bm 0
& \bm I & \bm 0  & \bm 0 & \bm A(\hat{\bm q}) &\bm 0\\
\bm I - [\delta \hat{\bm\mu}_v \times ]  &\bm 0  &\bm 0 &\bm 0  & \bm 0& \bm 0 & \bm I + [\delta \hat{\bm q}_v \times ]
\end{bmatrix}.
\end{equation}

\subsection{Adaptive Constrained EKF}
Define $\delta \hat{\bm x}_k^-$  and $\delta \hat{\bm x}_k^+$ as the {\em a prioir} and {\em a posteriori} estimates of the state vector at time $t_k$ and their corresponding estimation errors $\delta \tilde{\bm x}_k^-= \delta \bm x_k - \delta \hat{\bm x}_k^-$ and $\delta \tilde{\bm x}_k^+=\delta \bm x_k - \delta  \hat{\bm x}_k^+$ with associated covariances  $\bm P_k^-=E[\delta \tilde{\bm x}_k^- \delta \tilde{\bm x}_k^{-T}]$ and $\bm P_k^+=E[\delta \tilde{\bm x}_k^+ \delta \tilde{\bm x}_k^{+T}]$, where $E[\cdot]$ is the expected operator. The estimation update is
\begin{equation} \label{eq:innovation}
\delta \hat{\bm x}_k^+ = \delta \hat{\bm x}_k^- + \bm K_k (\bm z_k - \bm h(\delta \hat{\bm x}_k^-))
\end{equation}
The Kalman filter gain  minimizes the performance index $E(\| \delta \tilde{\bm x}_k^+ \|^2)=\mbox{tr}(\bm P_k^+)$ subject to the state constraints \eqref{eq:Dsigma<1}. Therefore, according to the Joseph formula, the constrained Kalman filter is the solution of the following optimization programming
\begin{align*}
\min_{\bm K_k}  \mbox{tr} & \big((\bm I - \bm K_k \bm H_k) \bm P_k^- (\bm I + \bm K_k \bm H_k)^T + \bm K_k \bm R_k \bm K_k^T  \big) \\
\mbox{subject to:} & \quad \bm D \bm\hat{\bm\sigma}_k^+ \prec \bm 1
\end{align*}
The gain projection technique can be applied to impose the inequality constraints for the estimation process \cite{Gupta-Hauser-2007,Teixeira-Chandrasekar-2008}. In this method if the {\em aprioir} estimate  satisfies the constraints but the unconstrained {\em aposteriori} estimate $\delta \hat{\bm x}_k^+$ does not satisfy them, then  the latter can be projected in the direction of the former until it reaches the constraint boundary. This effectively gives modified Kalman gain as follow
\begin{equation}
\bm K_k = \bm\beta_k \bm K^u_k,
\end{equation}
Here, $\bm K^u$ is the standard unconstrained Kalman gain and $\bm\beta_k=\mbox{diag}(1, 1, \cdots, \beta_{1_k}, \beta_{2_k}, \cdots, 1,1)$ where
\begin{equation} \notag
\beta_{i_k} =\left\{ \begin{array}{ll}
\mbox{sgn}(\bm k_{i_k}^T \bm e_k) -\frac{\hat{\sigma}_{i_k}^-}{\bm k_{i_k}^T \bm e} & \quad \mbox{if} \quad |\bm k_{i_k}^T \bm e_k| >1 \\
1 & \quad \mbox{otherwise} \end{array} \right.  \quad i=1,2
\end{equation}
with $\bm k^T_{1_k}$ and $\bm k^T_{2_k}$ being the last two row vectors of the unconstrained gain matrix. The unconstrained Kalman gain is given by the standard formula
\begin{subequations}
\begin{align}
\bm K_k^u & = {\bm P}_k^- \bm H_k^T {\bm S}_k^{-1} \\ \label{eq:Sk}
\bm S_k &= \bm H_k \bm P_k^- \bm H_k^T + \bm R_k .
\end{align}
Subsequently the state covariance matrix is updated from
\begin{equation}
{\bm P}_k^+ = \big( \bm I - \bm K_k \bm H_k  \big) {\bm P}_k^-
\end{equation}
\end{subequations}
followed by the propagation of the state and covariance matrix
\begin{subequations} \label{eq:KF_propagate}
\begin{align}\label{eq:state-prop}
\hat{\bm x}_{k+1}^- & = \hat{\bm x}_{k}^+ +
\int_{t_k}^{t_{k}+t_{\Delta}} \bm f(\bm x, \bm
0)\,{\text d} \tau\\ \label{eq:cov-prop}
{\bm P}_{k+1}^-&= \bm\Phi_{k} \bm P_{k}^+ {\bm\Phi}_{k}^T + \bm Q_{k}
\end{align}
\end{subequations}
The covariance matrices $\bm Q_{k}$ and $\bm R_{k}$ in expressions \eqref{eq:cov-prop} and \eqref{eq:Sk} are  associated with process noise and vision sensor noise. A priori knowledge of the  measurement noise covariance is not usually available because of unexpected noise and disturbance associated with 3D vision systems (such as  laser range sensors). This covariance matrix  can be readjusted in real-time from averaging the sequence of either the innovation matrix or the residual matrix \cite{Mehra-1970,Wang-2000,Gao-Wei-Zhong-Subic-2015}. Consider the residual sequence
\begin{equation} \notag
{\bm e}_k = \bm z_k - \bm H_k \delta \hat{\bm x}_k^+,
\end{equation}
which is the difference between pose measurement and the pose calculated from  {\em a priori} state. Subsequently, the associated  {\em innovation covariance matrix} can be estimated from averaging inside the moving window of size $w$, which can be empirically selected, i.e.,
\begin{equation} \label{eq:S_batch2}
{\bm\Sigma}_k  = \frac{1}{w}\sum_{i=1}^{w} {\bm e}_{k-i}
{\bm e}_{k-i}^{T}
\end{equation}
Then, the estimated observation covariance  is related to innovation covariance matrix \cite{Wang-2000} by
\begin{equation} \label{eq:R_resudua2}
\hat{\bm R_k} \approx {\bm\Sigma}_k + \bm H_k {\bm P}_k^+ \bm H_k^T
\end{equation}
Alternatively, the innovation covariance matrix \eqref{eq:S_batch2} can be recursively updated by
\begin{equation} \label{eq:S_recursive}
{\bm\Sigma}_{k+1} = \left\{ \begin{array}{ll} \frac{k-1}{k} {\bm\Sigma}_{k} + \frac{1}{k} \bm e_{k}  \bm e_{k}^T & \quad \text{if} \quad k<w\\
{\bm\Sigma}_{k} + \frac{1}{w} \Big( \bm e_{k} \bm e_{k}^T -
\bm e_{k-w}\bm e_{k-w}^T \Big) & \quad \text{otherwise}
\end{array} \right.
\end{equation}

\subsection{Observability Analysis}

\begin{figure}
\psfrag{time}[c][c][.8]{Time~(s)}
\psfrag{gramian}[c][c][.8]{cond$(\bm W_O)$}
\centering
\includegraphics[width=9.5cm]{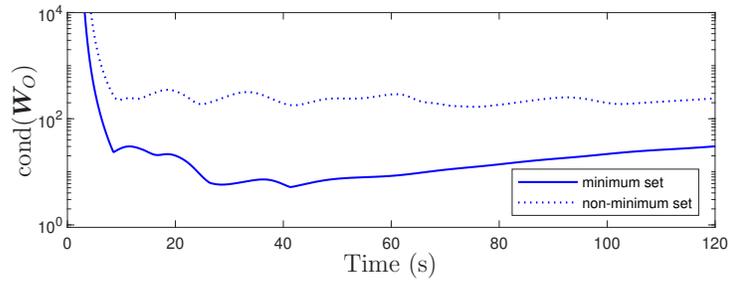} \caption{The time-history of the condition number of the Observability Gramian matrices associated with the minimum-set and the non-minimum set inertia parameters.} \label{fig:gramian}
\end{figure}

Only when a system is observable, then its states can be uniquely determined from the measurements regardless of the initial conditions. However, if a system is not fully observable, the  quality of state estimation may be adversely affected by the initial values or the the stability and convergence of the estimator may not be ascertained. A time-varying system is considered to be fully observable if its Observability Gramian Matrix (OGM) is not ill-conditioned \cite{Krener-Ide-2009,Yu-Cui-Zhu-2014}. Thus the degree of the observability of a system can be assessed by examining the condition number of its  OGM. In other words, the condition number of OGM reveals the degree of the observability  of the EKF estimator using the modeling technique. The discrete Observability Gramian matrix can be computed by:
\begin{equation}
\bm W_{O_k} = \sum_{j=1}^k \bm\Phi_{j/0}^T \bm H_j^T \bm H_j \bm\Phi_{j/0},
\end{equation}
where $\bm\Phi_{j/0}= \prod_{i=2}^j \bm\Phi_{i/i-1}$.  The above equation can be also equivalently written in the following recursive form
\begin{align} \notag
\bm W_{O_k} &= \bm W_{O_{k-1}} + \bm\Phi_{k/0}^T \bm H_k^T \bm H_k \bm\Phi_{k/0} \\
 \bm\Phi_{k/0} &= \bm\Phi_{k} \bm\Phi_{k-1/0}
\end{align}
The rank of ill-conditioning of the OG matrix can be quantitatively  represented by its condition number, i.e.,  a very large condition number means the matrix is ill-conditioned and thus not full-rank. The plots in Fig.~\ref{fig:gramian} show the time-varying condition numbers of two OG matrices corresponding to the  observation equations derived in terms of minimum set and non-minimum set inertia parameters. As expected,  the conditional number of the OGM of the estimator system formulated in terms of the minimum set inertia parameters   is significantly lower compared to the case of  non-minimum set inertia parameters. These results clearly demonstrate  a higher degree of observability of the estimator system formulated in terms of minimum set inertia parameters .

\section{Fault-Tolerant Vision Data Registration Using Motion Prediction}\label{sec:3D-registration}

The ICP registration is an iterative process that estimates the pose in two steps: i) Find the corresponding points on the model set assuming a coarse-alignment pose is given; ii) Resolve a fine-alignment pose corresponding to the two data sets by minimizing the sum of squared distances. Suppose $\{\bm q^{(n)}, \bm\rho^{(n)} \}$ represent rigid transformation at $n$th cycle of the ICP iteration and that the initial transformation $\{\bm q_k^{(0)}, \bm\rho_k^{(0)} \}$ is obtained from prediction of the states at the propagation step by:
\begin{equation} \label{eq:initial_pose}
\left\{ \begin{array}{l}
{\bm\eta}_k^{(0)} = \hat{\bm\mu}_k^- \otimes \hat{\bm q}_k^- \\
{\bm\rho}_k^{(0)} = \hat{\bm\rho}_{o_k}^- + \bm A(\hat{\bm q}_k^-) \hat{\bm\varrho}_k
\end{array} \right.
\end{equation}
Then, the problem of finding the correspondence between the two sets at the $n$th iteration can be formally expressed by
\begin{equation} \label{eq:ci}
\bm d_{i_k}^{(n)} = \mbox{arg} \min_{\bm d_j \in {\cal M} } \|\bm A \big({\bm\eta}_k^{(n)} \big) \bm c_{i_k}
+ {\bm\rho}_k^{(n)} - \bm d_j^{(n)} \| \quad \forall i=1,\cdots,m,
\end{equation}
and subsequently set ${\cal D}_k^{(n)}$ is formed. Now, we have two independent sets of 3D points ${\cal C}_k$  and ${\cal
D}_k^{(n)}$ both of which corresponds to the same shape but they may not completely coincide due to a rigid-body transformation. The next problem involves finding  the transformation represented by fine alignment $\{ {\bm\rho}_k^{(n+1)}, \; {\bm\eta}_k^{(n+1)} \}$ which
minimizes the distance between these two data sets~\cite{Besl-Mckay-1992}. That is  $ \forall \bm d_{i_k}^{(n)} \in {\cal D}_k^{(n)}, \bm c_{i_k}
\in {\cal C}_k$, we have
\begin{align} \label{eq:ICP}
& \varepsilon_k^{(n)} =  \min_{{\bm\eta}_k^{(n+1)}, {\bm\rho}_k^{(n+1)}} \frac{1}{m} \sum_{i=1}^m \| \bm A({\bm\eta}_k^{(n+1)}) \bm c_{i_k} +
{\bm\rho}_k^{(n+1)} - \bm d_{i_k}^{(n)} \|^2 \\ \notag & \qquad \mbox{subject to} \quad  \| \bar{\bm\eta}_k^{(n+1)} \| =1
\end{align}
where residual $\varepsilon_k^{(n)}$ is the metric fit error at the $n$-th iteration \cite{Horn-1987,Besl-Mckay-1992}.
To solve the above least-squares
minimization problem, we define the
cross-covariance matrix of the sets ${\cal C}_k$ and ${\cal D}_k^{(n)}$
by
\begin{equation}
\bm N^{(n)} = \mbox{cov}({\cal C}_k, {\cal D}_k^{(n)}) = \frac{1}{m} \sum_{i} \bm
c_{i_k} \big(\bm d_{i_k}^{(n)}\big)^T - {\bm c}_{o_k} \big({\bm d}_{o_k}^{(n)} \big)^T,
\end{equation}
where ${\bm c}_{o_k} = \frac{1}{m} \sum_{i} \bm c_{i_k}$ and ${\bm d}_{o_k}^{(n)}
= \frac{1}{m} \sum_{i} \bm d_{i_k}^{(n)}$ are the corresponding centroids of the point cloud data set. Let us also define the following symmetric
matrix constructed from  $\bm N$
\begin{equation} \notag
\bm M^{(n)} = \begin{bmatrix}
\mbox{tr}(\bm N^{(n)}) & \big(\bm n^{(n)} \big)^T \\
\bm n^{(n)} & \bm N^{(n)} + \big(\bm N^{(n)}\big)^T - \mbox{tr}\big(\bm N^{(n)} \big) \bm I
\end{bmatrix},
\end{equation}
where $\bm n^{(n)}=[N_{23}^{(n)} - N_{32}^{(n)}, N_{31}^{(n)} -N_{13}^{(n)}, N_{12}^{(n)}-N_{21}^{(n)}]^T$. Then, it
has shown in  \cite{Horn-1987} that minimization problem \eqref{eq:ICP} can be equivalently transcribed by the following quadratic programming
\begin{equation} \label{eq:quadratic}
\max_{\| {\bm\eta}_k^{(n+1)} \| =1} \big({\bm\eta}_k^{(n+1)} \big)^T \bm M^{(n)} \big( {\bm\eta}_k^{(n+1)} \big),
\end{equation}
which has the following closed-form
\begin{equation} \label{eq:q_rime}
\begin{array}{cc}
{\bm\eta}_k^{(n+1)} &={\mbox{eigenvector} \big(\bm M^{(n)} \big)}\\
& \lambda_{\rm max} \big(\bm M^{(n)} \big)
\end{array}
\end{equation}
Next, we can proceed with computation of the translation by
\begin{equation}  \label{eq:r_rime}
{\bm\rho}_k^{(n+1)} = {\bm d}_{o_k}^{(n)} - \bm A({\bm\eta}_k^{(n+1)}) {\bm c}_{o_k}
\end{equation}
The fine alignment obtained from  \eqref{eq:q_rime} and \eqref{eq:r_rime} can be used in Step I and then  continue the iterations until the residual error $\varepsilon_n^{(n)}$ becomes less
than the pre-specified  threshold $\varepsilon_{\rm th}$. That is
\begin{equation} \notag
\left\{ \begin{array}{ll}
\mbox{if} \quad \varepsilon_k^{(n)} < \varepsilon_{\rm th}   \quad & \bar{\bm\eta}_k = \bm\eta_k^{(n+1)}, \quad \bar{\bm\rho}_k = \bm\rho_k^{(n+1)}, \\
& \mbox{and exit the iteration}\\
\mbox{otherwise} & \mbox{go to \eqref{eq:ci} and use $\bm\eta^{(n+1)}$ and $\bm\rho_k^{(n+1)}$} \\
&  \mbox{as the new coarse alignment}
\end{array} \right.
\end{equation}
Suppose $n_{\rm max}$  represent the maximum number of iterations specified by a user. Then, the ICP loop is considered convergent if the matching error $\varepsilon_k^{(n)}$ is less than $\varepsilon_{\rm th}$  and $n \leq n_{\rm max}$. Otherwise, the ICP iteration at epoch $k$ is considered a failure. We should incorporate the refinement  rigid-body transformation  in the KF innovation sequence only when the ICP becomes convergent. To this goal, we introduce the following flagged variable to indicate wether the registration process at epoch $k$  is healthy or faulty
\begin{equation}
\gamma_k = \left\{ \begin{array}{ll} 1  \quad & \mbox{if} \quad  \varepsilon_k^{(n)} < \varepsilon_{\rm th} \quad \wedge \quad n \leq n_{\rm max} \\
0 & \mbox{otherwise} \end{array} \right.
\end{equation}
It is worth noting that although the necessary condition for convergent of ICP to a correct solution is small metric fit  error, the validity of the registration algorithm can be also examined by comparing the pose resolved by matching the date sets with the predicted pose obtained from  the dynamics model, i.e., $\bm\alpha_k =\bm z_k - \bm h(\delta \hat{\bm x}_k^-)$ calculated in  the innovation step \eqref{eq:innovation}. Then, the condition for detecting ICP fault ($\gamma_k=0$) can be described by
$\varepsilon \geq \varepsilon_{\rm th}$ and $\| \bm\alpha_k \|_W \geq \alpha_{\rm th}$, where $\alpha_{\rm th}$  is the pose error threshold, and $\| \cdot \|_W$ denotes weighted Euclidean norm. Here, the weight matrix can be properly selected as $\bm W= \mbox{diag}(\bm I , \; L \bm I  )$, where $L$ is the called the characteristic length which is empirically obtained as the ratio of the maximum linear and angular errors. Once ICP fault is detected, the pose-tracking fault recovery is rather straightforward. To this end, the Kalman filter gain is set to
\begin{equation} \label{eq:K_phi}
\bm K_k  = \gamma_k \bm\beta_k {\bm P}_k^- \bm H_k^T \hat{\bm S}_k^{-1}
\end{equation}
Clearly $\bm K_k = \bm 0$ when ICP fault is detected, in which case the observation information is not incorporated in the estimation process  for updating the state and the estimator covariance, i.e., $\gamma_k=0 \; \Longrightarrow \; \bm K_k = 0$ and thus $\hat{\bm x}_k^+ = {\bm x}_k^-$ and $\bm P_k^+ = \bm P_k^-$. In other words, the estimator relies on the dynamics model for pose estimation until ICP becomes convergent again.

%========================================================
\section{Optimal Rendezvous \& Guidance for Visual Servoing} \label{sec:guidance}
%========================================================
This section presents an optimal robot guidance approach for rendezvous and smooth interception of a moving object based on visual feedback. As schematically illustrated in Fig.~\ref{fig:robot_target}, the position of the end-effector and the grasping point are denoted by  $\bm r$ and $\bm\rho$, respectively.
The end-effector and the grasping point are expected to arrive at a rendezvous-point
simultaneously with the same velocity in order to avoid impact at
the time of grasping. Suppose the optimal trajectory is manifested  by the
following second order system
\begin{equation} \label{eq:sys_xr}
\ddot{\bm r} =\bm u,
\end{equation}
which can be formally rewritten as $\dot{\bm\xi} =[ \dot{\bm r}^T  \; \bm u^T ]^T$ with state vector $\bm\xi^T=[\bm r^T \; \dot{\bm r}^T]$, and terminal condition $\bm r(t_f) =\bm\rho(t_f)$ and $\dot{\bm r}(t_f) =\dot{\bm\rho}(t_f)$ at terminal time $t_f$. In the following analysis, we seek a time-optimal solution to the input $\bm u$ subject to the acceleration limit $\|\ddot{\bm r} \|\leq a_{\rm max}$ and the aforementioned terminal constraints, i.e.,
\begin{subequations}
\begin{align}
\mbox{minimize}  &  \qquad\int_t^{t_f} 1 \; d \tau\\ \label{eq:a_max}
\mbox{subject to:}   & \qquad \| \bm u(\tau) \| \leq a_{\rm max}  \qquad t \leq \tau \leq t_f \\ \label{eq:terminal}
& \qquad \bm r(t_f) = \bm\rho(t_f), \qquad \dot{\bm r}(t_f) = \dot{\bm\rho}(t_f)
\end{align}
\end{subequations}
Defining the vector of Lagrangian multiplier as $\bm\lambda$, one can write the expression of the system Hamiltonian as
\[ H = 1 + \bm\lambda^T \dot{\bm\xi} \]
According to the optimal control
theory~\cite{Anderson-Moore-1990}, the costate  must satisfy
\begin{equation} \notag
\dot{\bm\lambda} = -\frac{\partial H}{\partial \bm\xi} \quad \mbox{hence} \quad \bm\lambda^* = \begin{bmatrix} \bm c_1 \\ - \bm c_1 \tau + \bm c_2 \end{bmatrix} \quad \forall \tau\in[t, \; t_f],
\end{equation}
where $^*$ indicates optimal values, vectors $\bm c_1$ and $\bm c_2$ are unknown constant vectors to be found from the boundary conditions \eqref{eq:terminal}. Thus
\begin{equation} \label{eq:H*}
H(\bm\xi^*, \bm\lambda^*, \bm u) = 1 + \bm c_1^T \dot{\bm r} + (-\bm c_1 \tau + \bm c_2 )^T \bm u.
\end{equation}
The Pontryagin's principle dictates that the optimal input $\bm u^*$ satisfies
\begin{equation} \notag
\bm u^* = \mbox{arg} \min_{\bm u} H(\bm\xi^*, \bm\lambda^*, \bm u).
\end{equation}
Defining vector $\bm p = -\bm c_1 \tau + \bm c_2 $, one can infer from \eqref{eq:H*} that the optimal control input $\bm u^*$ minimizing the Hamiltonian should be aligned with unit vector $-\bm p/\| \bm p \|$. Therefore, in view of the acceleration limit constraint \eqref{eq:a_max}, the optimal input must take the following structure
\begin{equation} \label{eq:u}
\bm u^*  =-\frac{-\bm c_1 \tau  + \bm c_2}{\| -\bm c_1 \tau  + \bm c_2 \|} a_{\rm max} \quad \forall \tau\in[t, \; t_f]
\end{equation}
However, the optimal terminal time $t_f$ along with vectors $\bm c_1$ and $\bm c_2$ remain to be found. The
{\em optimal Hamiltonian} calculated at optimal point $\bm u^*$ and $\bm\xi^*$
must also satisfy
\[ \frac{d}{d \tau} H(\tau) =0 \]
and hence
\begin{equation} \notag
1+ \bm c_1^T \dot{\bm r} + \|  \bm c_1 \tau - \bm c_2 \| a_{\rm max} = \mbox{const}.
\end{equation}
Applying the initial and final conditions  to the above equation yields
\begin{align} \notag
\Delta  H(\bm c_1, \bm c_2, t_f) &=  \big(\| \bm c_1 t - \bm c_2 \| -  \| \bm c_1 t_f - \bm c_2 \| \big) a_{\rm max} \\ \label{eq:H=0} & +  \bm c_1^T \big(\dot{\bm\rho}(t_f) - \dot{\bm r}(t_f) \big) =0.
\end{align}
Finally applying the terminal conditions \eqref{eq:terminal} to \eqref{eq:u} and combining the resultant equations with \eqref{eq:H=0}, we arrive at the following error equation in terms of seven unknowns $\bm \chi^T = [\bm c_1^T \;  \bm c_2^T \; t_f ]$.
\begin{equation} \label{eq:phi}
e(\bm \chi) = \left\| \begin{bmatrix} \int_{t}^{t_f} \bm u(\bm c_1, \bm c_2, \nu) d \nu - \dot{\bm\rho}(t_f) \\ \int_{t}^{t_f} \int_{t}^{\mu} \bm u(\bm c_1, \bm c_2, \nu) d \nu d \mu  - \bm\rho(t_f) \\ \Delta  H(\bm c_1, \bm c_2, t_f) \end{bmatrix} \right\| = 0,
\end{equation}
where $\bm\rho(t_f)$ and $\dot{\bm\rho}(t_f)$ are the position and velocity along the target trajectories at the interception, which can be calculated from
\begin{align} \notag
\dot{\bm\rho}(t_f) &=   \int_t^{t_f}  \bm A({\bm q})[ {\bm\omega} \times( {\bm\omega} \times \hat{\bm\varrho}) + \bm\phi({\bm\omega}, \hat{\bm\sigma})\times \hat{\bm\varrho} \; ] d \tau \\
\bm\rho(t_f) &= \int_t^{t_f} \dot{\bm\rho}(\tau) d \tau.
\end{align}
The set of seven nonlinear equations in \eqref{eq:phi} can be solved for seven unknowns $\{\bm c_1, \bm c_2, t_f \}$ by a numerical method. Then the solution can be substituted
into \eqref{eq:u} to resolve the optimal robot trajectories by integration. Clearly, the error $e$ computed at correct variable should vanishes at the terminal time. Therefore, one may call an unconstrained optimization function  (\verb"fminunc" function in Matlab) to find the solution based on a quasi-Newton method \cite{Boyarko-Yakimenko-2011}.

%------------------------------------------------------
\section{End-to-End Experimental Validation} \label{sec:experiment}
%------------------------------------------------------

This section presents results obtained from a hardware-in-loop simulation demonstrating the performance of the adaptive fault-tolerant vision guided robotic system. As described in \cite{Aghili-2022}, the experimental setup consists of servicing and simulator manipulators, a Laser Camera System (LCS), and a  target which is scaled model of a micro-satellite. The entire setup is enclosed within a shroud covered by black curtains, which have ability to absorb the laser beam and hence minimizing outliers in the 3-D imaging data. The LCS unit employed in this experiment is from Neptec Design Group Ltd, which flew successfully onboard space shuttle Discovery during mission STS-105 to the International Space Station (ISS) and subsequently generated real-time 3-D imaging data  \cite{Samson-English-Deslauriers-Christie-2004}.
%The Neptec's Laser LCS is preferable because of its robustness in the face of the harsh lighting conditions of space.  The immunity  against  solar  illumination  comes  from the choice of a 1500 nm wavelength laser source which corresponds  to  a minimum overlap in the solar spectrum  \cite{Samson-English-Deslauriers-Christie-2004}. 
The target is mounted on the robot simulator, which simulates the relative motions between the chaser satellite and the target satellite in orbit through a high fidelity simulator.
The orbital force/torque perturbations are simulated by random variables with covariances
$E[\bm\epsilon_f \bm\epsilon_f^T] = 2 \times 10^{-6} \bm I~\mbox{m}^2/\mbox{s}^4$ and
$E[\bm\epsilon_{\tau} \bm\epsilon_{\tau}^T] =3 \times 10^{-5} \bm I~\mbox{rad}^2/\mbox{s}^4$. The other  robotic arm is controlled by  the proposed fault-tolerant vision guided system to autonomously approach the mockup and capture its grapple fixture. The LCS feeds the image registration algorithm with range data at a rate of 2~Hz, the pre-capture trajectory planter is updated at 1 Hz based on latest estimation of the target's inertial parameters and states, and the control servo loop runs  at 1 kHz  \cite{Aghili-Nmavar-2006}. It should be pointed out that the update rate of the LCS used in our setup  is sufficient to track our target which movies at relatively low speed. However, the vision guided robotic system may not be able to operate properly for a faster moving target unless a laser scanner with higher update rate is utilized to feed the visual servoing system. The robotic arm carrying the satellite mockup replicates the dynamics of a free-floater with the following parameters: $I_{xx}=14~\text{kgm}^2$,
$I_{yy}=10~\text{kgm}^2$, $I_{zz}=6~\text{kgm}^2$, and $\bm\varrho=[-0.15 \; 0.03 \; -0.05]^T$~m. Hence the dimensionless inertia parameters are: $\sigma_1=0.2857$ and $\sigma_2=-0.80$. These values are used to determine the manipulator-driven motion of the target as  free-floating object with initial drift velocity
$\dot{\bm\rho_o}(0)=[6 \; -4 \; 5]^T~\text{mm/s}$ and initial angular velocity
$\bm\omega(0)=[0.15 \; -0.18 \; -0.12]~\text{rad/s}$.

\begin{figure}
\psfrag{eth}[c][c][.7]{$\varepsilon_{\rm th}$}
\psfrag{ICP error}[c][c][.7]{metric fit error $\varepsilon$}
\centering
\includegraphics[width=9.5cm]{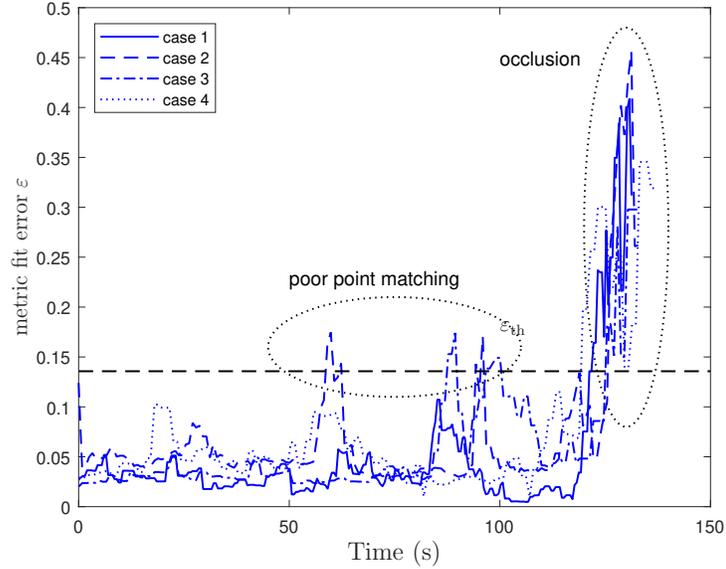} \caption{Time-histories of the ICP metric fit  error.} \label{fig:icp_error}
\end{figure}

\begin{figure}
\psfrag{time}[c][c][.8]{Time~(s)}
\psfrag{rho}[c][c][.7]{$\bar{\bm\rho}$~(m)}
\psfrag{eta}[c][c][.7]{Attitude $\bar{\bm\eta}$}
\psfrag{et1}[l][l][.6]{$\eta_1$}\psfrag{et2}[l][l][.6]{$\eta_2$}\psfrag{et3}[l][l][.6]{$\eta_3$}\psfrag{et4}[l][l][.6]{$\eta_4$}
\psfrag{t3}[c][c][.8]{$t_3$}
\centering
\includegraphics[width=9.5cm]{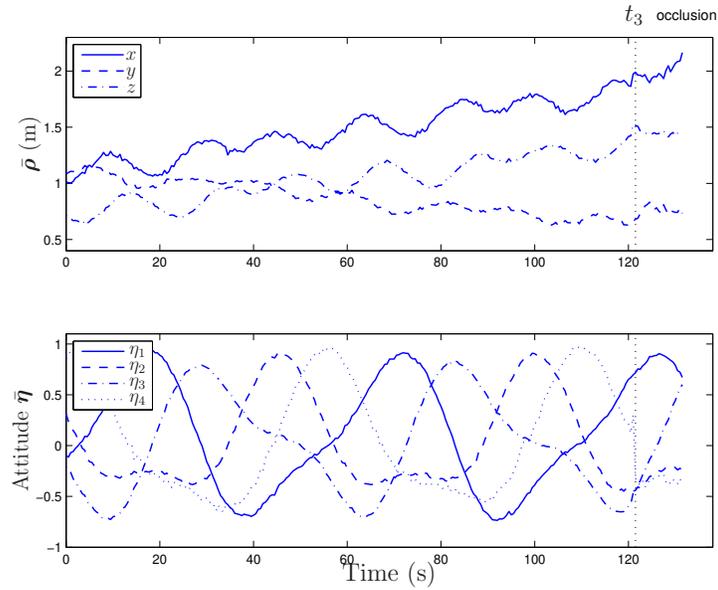} \caption{Raw ICP pose estimation.} \label{fig:icp_pose}
\end{figure}

\begin{figure}
\psfrag{time}[c][c][.8]{Time~(s)}
\psfrag{omega}[c][c][.7]{$\hat{\bm\omega}$~(rad/s)}
\psfrag{dot-rho-o}[c][c][.7]{$\hat{\dot{\bm\rho}}_o$}
\psfrag{sigma}[c][c][.7]{$\hat{\bm\sigma}$}
\psfrag{vrho}[c][c][.7]{$\hat{\bm\varrho}$~(m)}
\psfrag{s1}[l][l][.6]{$\sigma_1$}
\psfrag{s2}[l][l][.6]{$\sigma_2$}
\psfrag{gam0}[l][l][.6]{$\gamma=0$}
\psfrag{gam1}[l][l][.6]{$\gamma=1$}
\centering
\includegraphics[width=8.5cm]{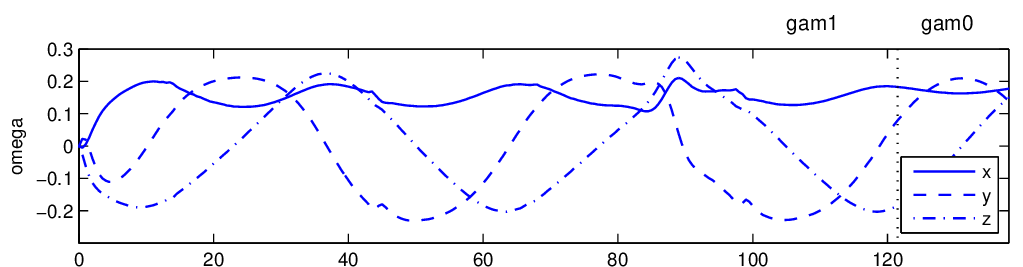}\\
\vspace*{4mm}
\includegraphics[width=9.5cm]{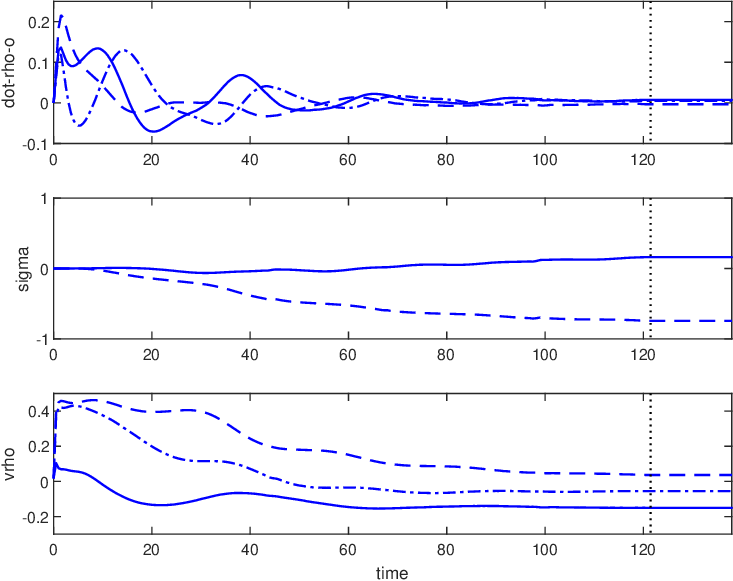}
\caption{Motion and parameter estimation of the target, test case 1.} \label{fig:motion_estimation}
\end{figure}

\begin{figure}
\psfrag{dot-rho-o}[c][c][.7]{$\hat{\dot{\bm\rho}}_o$}
\psfrag{sigma}[c][c][.7]{$\hat{\bm\sigma}$}
\psfrag{vrho}[c][c][.7]{$\hat{\bm\varrho}$~(m)}
\centering
\includegraphics[width=9.5cm]{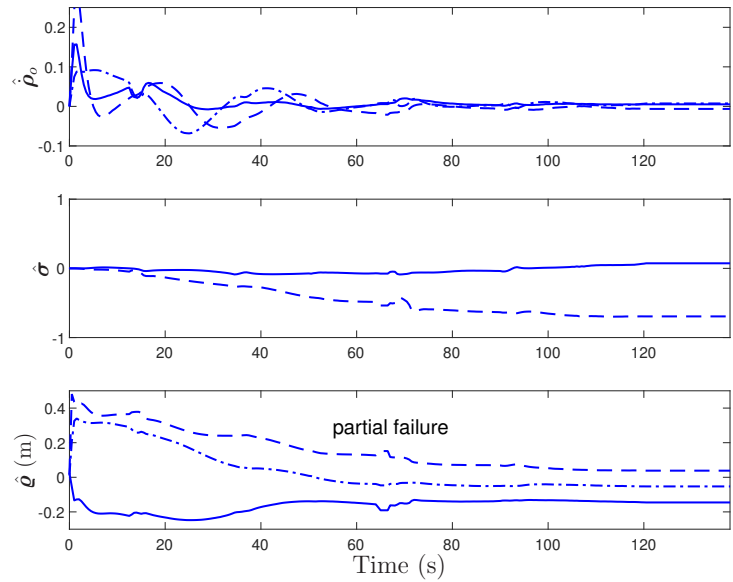}
\caption{Parameter estimation of the target, test case 2.} \label{fig:motion_estimation2}
\end{figure}

\begin{figure}
\centering
\psfrag{rho}[c][c][.7]{$\bm\rho$~(m)} \psfrag{t3}[c][c][.8]{$t_3$}
\includegraphics[width=9.5cm]{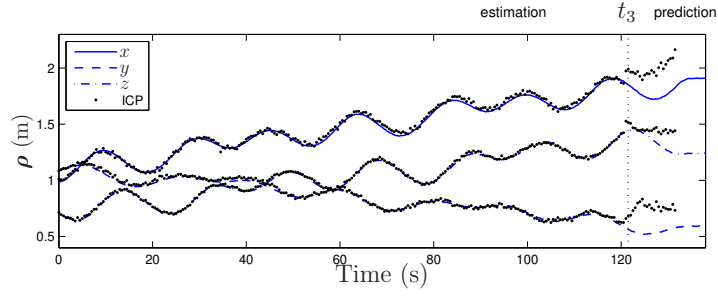} \caption{The estimation and prediction position of the grapple-fixture.} \label{fig:predicted_rho}
\end{figure}

\begin{figure}
\psfrag{error}[c][c][.7]{$\| \bm\rho - \hat{\bm\rho} \|$~(m)}
\centering
\includegraphics[width=9.5cm]{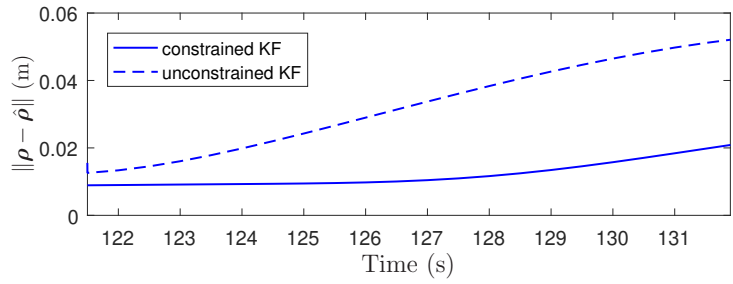} \caption{Trajectories of predicted position errors after the occlusion obtained from the unconstrained and constrained KF.} \label{fig:prediction_err}
\end{figure}

\begin{figure}
\psfrag{distance}[c][c][.7]{$\bm\rho, \; \bm r$~(m)}
\psfrag{linear velocity}[c][c][.7]{$\dot{\bm\rho}, \; \dot{\bm r}$~(m/s)}
\psfrag{t0}[c][c][.8]{$t_0$} \psfrag{t1}[c][c][.8]{$t_1$} \psfrag{t2}[c][c][.8]{$t_2$} \psfrag{t3}[c][c][.8]{$t_3$} \psfrag{tf}[c][c][.8]{$t_f$}
\centering
\includegraphics[width=9.5cm]{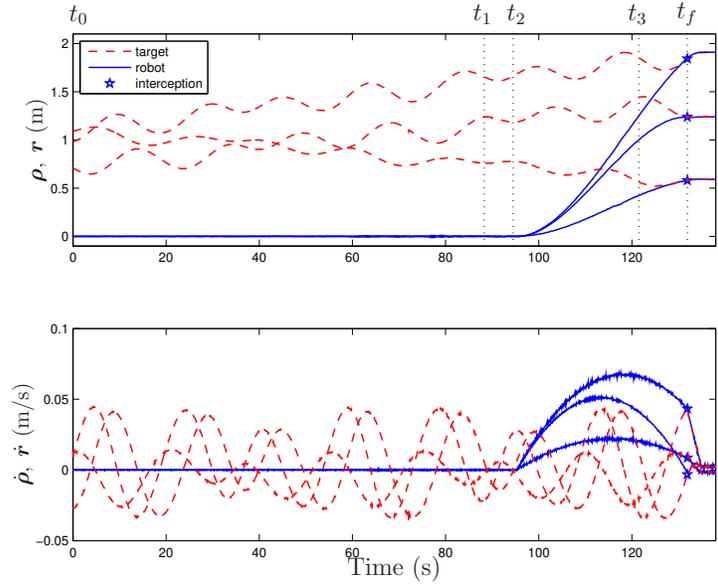} \caption{Trajectories of the robot end-effector and the target's grapple-fixture.} \label{fig:robot_traj}
\end{figure}

\begin{figure}
\psfrag{relative distance}[c][c][.7]{relative distance (m)}
\psfrag{relative velocity}[c][c][.7]{relative velocity  (m/s)}
\centering
\includegraphics[width=9.5cm]{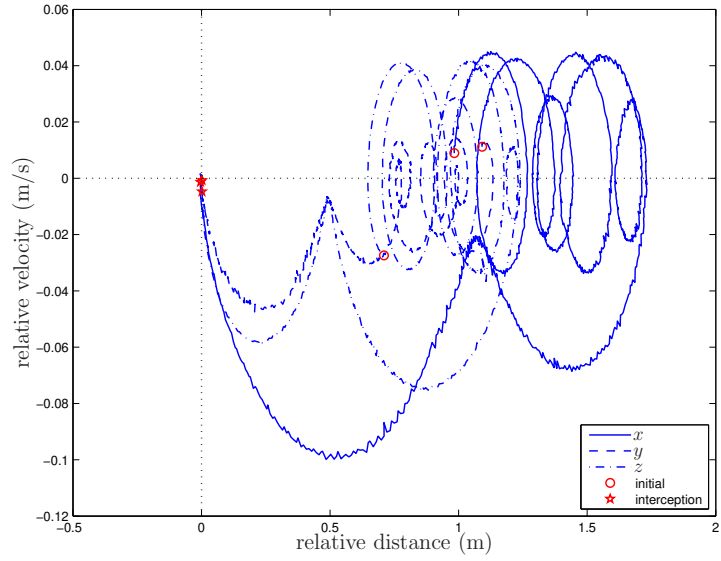} \caption{The relative velocity versus the relative distance.} \label{fig:phase_plane}
\end{figure}

\begin{figure}
\psfrag{rel distance}[c][c][.7]{$\| \bm r - \bm\rho\|$ (m)}
\psfrag{rel velocity}[c][c][.7]{$\| \dot{\bm r} - \dot{\bm\rho} \|$ (m/s)}
\centering
\includegraphics[width=9.5cm]{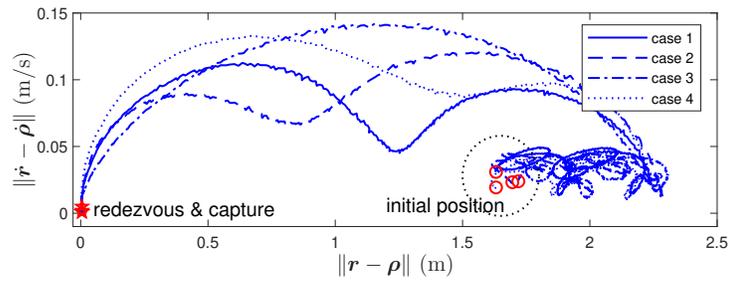} \caption{Rendezvous and capture trajectories of test cases.} \label{fig:test_cases}
\end{figure}

We conduced four test cases for autonomous rendezvous \& capture operations with different initial conditions.  The time-histories of the  ICP fit metric errors based on streaming 3D data associated with four test cases are shown in Fig.~\ref{fig:icp_error}. It is apparent from the plots that total failure of the visual  system due to occlusion always occurs few seconds prior to grasping in all four cases, e.g., failure of the vision system in the test case 1 occurs at time $t_2=121.5$~s. This is because the approaching robot inevitably comes into the field-of-view (FOV) of the vision sensor prior to grasping. Nevertheless,  partial failure  of the vision  system occasionally occurs, i.e., test cases 2 \& 3, that is attributed to poor point-matching performance of the image registration algorithm  in the presence of noises and outliers. The time-history of the position and orientation of the mockup satellite obtained from ICP algorithm are illustrated in Fig.~\ref{fig:icp_pose}. The corresponding trajectories of the estimated states and the parameters for the test cases involving partial and complete failure of visual feedback are shown in Fig.~\ref{fig:motion_estimation} and \ref{fig:motion_estimation2}. The motion planner automatically generates robot trajectory based on estimated states and parameters in conjunction with the optimal rendezvous and guidance scheme described in Section~\ref{sec:guidance} upon convergence of the dynamic estimator. As shown in Fig.~\ref{fig:motion_estimation}, the estimator converges at the time $t_1=88.3$~s while the convergence is detected by observing the magnitude of the state estimation covariance matrix. Subsequently, the robot departure time is set to $t_2=94.5$~s and the interception  is predicted to happen  at the time $t_f=131.9$~s based on prediction of the dynamic estimator. Trajectories of the predicted  position of the grapple fixture calculated from the latest estimation of the motion states and parameters of the target are shown in Fig.~\ref{fig:predicted_rho}.  It is apparent from the plots in the figure that the translational motion of target evolves more than 30 cm, which is much larger than 4 cm capture envelope of the robotic grasping hand, during the last 10 seconds when the visual feedback fails. A comparison of the trajectories of
the visual measurements  and those of the predicted position in Fig.~\ref{fig:predicted_rho} reaffirms that the visual measurements are not reliable at the final stages of the robot approach after time $t_3=121.5$~s due to the obstruction of the vision system. Nevertheless, the plots  clearly show that the adaptive KF is able to continue  providing accurate pose estimates in spite of the failure of the vision sensor. To this end,  the consistency and accuracy of the adaptive visual servoing using the constrained and unconstrained estimators are comparatively illustrated in Fig.~\ref{fig:prediction_err}. It is apparent from the plots that incorporation of  unconstrained estimator in the adaptive visual servo gives rise to the position prediction error when visual occlusion occurs. Position and velocity trajectories associated with the robot's end-effector and the satellite grapple fixture are shown in Fig.~\ref{fig:robot_traj}. Critical times are annotated in the figure to illustrate the sequence of events as follows: the estimator converges at the time $t_1=88.3$~s, the robot departure time is set to $t_2=94.5$~s, the vision system is occluded by the approaching arm at time $t_2=121.5$~s, and finally the interception  successfully  occurs  at the time $t_f=131.9$~s. The pose trajectories of the spacecraft mockup and its angular velocity are also obtained by making use of the manipulator kinematic model that is used as the ground truth for a comparison. It is evident that the robot intercepts the handle with the same velocity at the time of the rendezvous at  the predicted rendezvous time  and then the motion stopped by braking the robotic arms. Trajectories of the relative velocity versus relative position are also depicted in  Fig.~\ref{fig:phase_plane}, which clearly demonstrates smooth capture of the moving target despite the total failure of the vision system in last ten seconds of the grasping operation. Trajectories of the rendezvous \& capture for all test case are plotted in  Fig.~\ref{fig:test_cases} and the rendezvous position and velocity errors are given in Table~-\ref{tab:capture}.  The test results indicate  average position error of $2.4$~cm, which is lower than the 4 cm capture envelope of the robotic gripper hand used in our experiment, and  average rendezvous velocity of 5.8 mm/s, which is tantamount to 1.2 mili-Joule impact energy assuming 200 kg target. Thus the rendezvous \& capture operation has been  successful inspire of the occlusion. To this end, it is worth pointing out that from $2.4$~cm position prediction error during 10 about 10 occlusion, one can extrapolate the maximum occlusion time for a yet successful  rendezvous \& capture operation to be 16 seconds given the 4 cm capture envelope.

%=============================================================
\begin{table}
\caption{Rendezvous position and velocity errors.}
\begin{center}
\begin{tabular}{ccc}
\hline \hline
test case & position error & relative velocity \\
\hline
1 & 2.1 cm $< 4^*$ cm  & 4.2 mm/s  \\
2 & 2.5 cm $< 4^*$ cm  & 6.4 mm/s  \\
3 & 1.8 cm $< 4^*$ cm & 5.7 mm/s   \\
4 & 3.1 cm $< 4^*$ cm  & 7.1 mm/s   \\
\hline \hline
\end{tabular}\\
$*$ capture envelope
\end{center}\label{tab:capture}
\end{table}
%=============================================================

%---------------------------------------------------------------
%\begin{figure}[t]
%\centering{\includegraphics[clip,width=8.5cm]{cart_black}}
%\caption{The
%experimental setup} \label{fig:lcs_cart}
%\end{figure}
%---------------------------------------------------------------

\psfrag{time}[c][c][.8]{Time~(s)}
\psfrag{x}[c][c][.6]{$x$}
\psfrag{y}[c][c][.6]{$y$}
\psfrag{z}[c][c][.6]{$z$}

%=============================================================
%=============================================================

%=============================================================
%=============================================================

%=============================================================
%=============================================================

%=============================================================
%=============================================================

%=============================================================
%=============================================================

%=============================================================
%=============================================================

%=============================================================
%=============================================================

%=============================================================
%=============================================================

%=============================================================
%=============================================================

%=============================================================
%=============================================================

%=============================================================
%\begin{figure}
%\psfrag{sig1}[c][c][.6]{$\sigma_1$} \psfrag{sig2}[c][c][.6]{$\sigma_2$} \psfrag{sig3}[c][c][.6]{$\sigma_3$} %\psfrag{Gamma}[c][c][.7]{$\Gamma(\hat{\bm\sigma})$} \psfrag{sigma}[c][c][.7]{$\hat{\bm\sigma}$}
%\centering
%\includegraphics[width=8.5cm]{sigma_unconstrained} \caption{Time-histories of the parameter estimations.} %\label{fig:sigma_unconstrained}
%\end{figure}
%=============================================================

\section{Conclusions}

An adaptive vision-guided robotic system for on-orbit servicing has been presented, enabling precise and smooth capture of a moving object, even in cases where the vision system temporarily lost pose tracking. The hierarchical visual servoing framework was developed through the integrated use of a variant of ICP registration, a constrained noise-adaptive Kalman filter, fault detection logic, and optimal guidance and control. We demonstrated the existence of elegant equality and inequality constraints related to dimensionless inertial properties of free-floating objects and subsequently developed an innovative dynamics model using a minimal set of parameters. An adaptive constrained Kalman filter was then designed to ensure consistent parameter and state estimation by imposing the triangle inequality and dynamically adjusting the observation covariance matrix in real-time. Fault detection and pose recovery strategies were implemented based on prediction errors and ICP metric fit discrepancies, integrated with the closed-loop image registration process and dynamic estimator. Finally, an optimal rendezvous and guidance strategy was developed, utilizing processed visual feedback to execute robot trajectories that intercept the target's grasping point with zero relative velocity as quickly as possible, subject to acceleration constraints. Experimental results demonstrated the system’s ability to smoothly capture a free-floating object using a vision-guided robotic arm, even during a 10-second period where the vision system failed in the final approach and capture phase.

%-------------------------------------------------------------
\bibliographystyle{IEEEtran}
%\bibliography{C:/Users/farha/OneDrive/Documents/Publications/bib/references}
%\bibliography{references}

\begin{thebibliography}{10}

\bibitem{Aghili-2022}
F.~Aghili, ``Fault-tolerant and adaptive visual servoing for capturing moving
  objects,'' \emph{IEEE/ASME Trans. on Mechatronics}, vol.~27, no.~3, pp.
  1773--1783, 2022.

\bibitem{Inaba-2000}
N.~Inaba and M.~Oda, ``Autonomous satellite capture by a space robot: world
  first on-orbit experiment on a japanese robot satellite ets-vii,'' in
  \emph{Proceedings 2000 ICRA. Millennium Conference. IEEE International
  Conference on Robotics and Automation. Symposia Proceedings (Cat.
  No.00CH37065)}, vol.~2, 2000, pp. 1169--1174 vol.2.

\bibitem{Aghili-2023}
F.~Aghili, ``Autonomous sequential sub-maneuvers in pre- and post-grasping
  moving objects using obstructed 3-d vision data,'' \emph{IEEE Trans. on
  Aerospace and Electronic Systems}, June 2023.

\bibitem{Wenberg-2020}
D.~Wenberg, M.~Kutzer, L.~Devries, J.~Gregory, M.~Sanders, and J.~S. Kang,
  ``Development of on-orbit assembly demonstrator in 3u cubesat form factor,''
  in \emph{2020 IEEE Aerospace Conference}, 2020, pp. 1--11.

\bibitem{Aghili-2011k}
F.~{Aghili}, ``A prediction and motion-planning scheme for visually guided
  robotic capturing of free-floating tumbling objects with uncertain
  dynamics,'' \emph{IEEE Transactions on Robotics}, vol.~28, no.~3, pp.
  634--649, June 2012.

\bibitem{Wang-Meng-2020}
B.~Wang, Z.~Meng, C.~Jia, and P.~Huang, ``Reel-based tension control of
  tethered space robots,'' \emph{IEEE Transactions on Aerospace and Electronic
  Systems}, vol.~56, no.~4, pp. 3028--3043, 2020.

\bibitem{Aghili-Parsa-2007b}
F.~Aghili and K.~Parsa, ``Adaptive motion estimation of a tumbling satellite
  using laser-vision data with unknown noise characteristics,'' in \emph{2007
  IEEE/RSJ International Conference on Intelligent Robots and Systems}, Oct
  2007, pp. 839--846.

\bibitem{Aghili-p03}
F.~Aghili, ``Robust impedance-matching of manipulators interacting with unknown
  environments,'' US Patent, $7,688,016$.

\bibitem{Wang-Huang-2015}
D.~Wang, P.~Huang, and Z.~Meng, ``Coordinated stabilization of tumbling targets
  using tethered space manipulators,'' \emph{IEEE Transactions on Aerospace and
  Electronic Systems}, vol.~51, no.~3, pp. 2420--2432, July 2015.

\bibitem{Aghili-2013}
F.~Aghili, ``Pre- and post-grasping robot motion planning to capture and
  stabilize a tumbling/driftig free-floater with uncertain dynamics,'' in
  \emph{IEEE International Conf. on Robotics \& Automation}, Karlsruhe,
  Germany, May~6--10 2013, pp. 5441--5448.

\bibitem{Aghili-Dupuis-Piedboeuf-deCarufel-1999}
F.~Aghili, E.~Dupuis, J.-C. {Piedb{\oe}uf}, and J.~{de Carufel},
  ``Hardware-in-the-loop simulations of robots performing contact tasks,'' in
  \emph{International Symposium on Artificial Intelligence and Robotics \&
  Automation in Space: {i-SAIRAS}}, M.~Perry, Ed.\hskip 1em plus 0.5em minus
  0.4em\relax Noordwijk, The Netherland: ESA Publication Division, 1999, pp.
  583--588.

\bibitem{Aghili-Namvar-2008}
F.~Aghili and M.~Namvar, ``Scaling inertia properties of a manipulator payload
  for 0-g emulation of spacecraft,'' \emph{The International Journal of
  Robotics Research}, vol.~28, no.~7, pp. 883--894, July 2009.

\bibitem{Kang-Zhu-2021}
J.~Kang, Z.~H. Zhu, and L.~F. Santaguida, ``Analytical and experimental
  investigation of stabilizing rotating uncooperative target by tethered space
  tug,'' \emph{IEEE Transactions on Aerospace and Electronic Systems}, vol.~57,
  no.~4, pp. 2426--2437, 2021.

\bibitem{Aghili-Parsa-2009b}
F.~Aghili and K.~Parsa, ``A reconfigurable robot with lockable cylindrical
  joints,'' \emph{IEEE Trans. on Robotics}, vol.~25, no.~4, pp. 785--797,
  August 2009.

\bibitem{Aghili-2020a}
F.~Aghili, ``Optimal trajectories and robot control for detumbling a
  non-cooperative satellite,'' \emph{{AIAA} Journal of Guidance, Control, and
  Dynamics}, vol.~43, no.~10, pp. 1952--1959, 2020.

\bibitem{Mithun-2018}
P.~Mithun, H.~Pandya, A.~Gaud, S.~V. Shah, and K.~M. Krishna, ``Image based
  visual servoing for tumbling objects,'' in \emph{2018 IEEE/RSJ International
  Conference on Intelligent Robots and Systems (IROS)}, 2018, pp. 2901--2908.

\bibitem{Liang-2022}
C.~Liang, R.~Wang, and X.~Liu, ``A visual-based servo control method for space
  manipulator assisted docking,'' in \emph{2022 IEEE 6th Information Technology
  and Mechatronics Engineering Conference (ITOEC)}, vol.~6, 2022, pp. 72--76.

\bibitem{Mahmood-Vagvolgyi-2020}
A.~Mahmood, B.~P. Vagvolgyi, W.~Pryor, L.~L. Whitcomb, P.~Kazanzides, and
  S.~Leonard, ``Visual monitoring and servoing of a cutting blade during
  telerobotic satellite servicing,'' in \emph{2020 IEEE/RSJ International
  Conference on Intelligent Robots and Systems (IROS)}, 2020, pp. 1903--1908.

\bibitem{Aghili-Kuryllo-Okouneva-English-2010a}
F.~Aghili, M.~Kuryllo, G.~Okouneva, and C.~English, ``Fault-tolerant
  position/attitude estimation of free-floating space objects using a laser
  range sensor,'' \emph{IEEE Sensors Journal}, vol.~11, no.~1, pp. 176--185,
  Jan. 2011.

\bibitem{Howard-2008}
R.~Howard, A.~F. Heaton, R.~M. Pinson, C.~L. Carrington, J.~E. Lee, T.~C.
  Bryan, B.~A. Robertson, S.~H. Spencer, and J.~E. Johnson, ``The advanced
  video guidance sensor: Orbital express and the next generation,'' 2008.
  


\bibitem{Shang-Jasiobedzki-2005}
L.~Shang, P.~Jasiobedzki, and M.~Greenspan, ``Discrete pose space estimation to
  improve icp-based tracking,'' in \emph{3-D Digital Imaging and Modeling,
  2005. 3DIM 2005. Fifth International Conference on}, June 2005, pp. 523--530.

\bibitem{Flores-Abad-Ma-2013}
A.~Flores-Abad, O.~Ma, K.~Pham, and S.~Ulrich, ``A review of space robotics
  technologies for on-orbit servicing,'' \emph{Progress in Aerospace Sciences},
  vol.~68, pp. 1--26, 2013.

\bibitem{NASA-2010}
NASA, ``On-orbit satellite servicing staudy, project report,'' National
  Aeronautics and Space Adminstration (NASA), Goddard Space Flight Center,
  Tech. Rep., 2010.

\bibitem{Samson-English-Deslauriers-Christie-2004}
C.~Samson, C.~English, A.~Deslauriers, I.~Christie, F.~Blais, and F.~Ferrie,
  ``Neptec 3{D} laser camera system: From space mission {STS}-105 to
  terrestrial applications,'' \emph{Canadian Aeronautics and Space Journal},
  vol.~50, no.~2, pp. 115--123, 2004.

\bibitem{Aghili-Kuryllo-Okouneva-English-2010b}
F.~Aghili, M.~Kuryllo, G.~Okouneva, and C.~English, ``Robust vision-based pose
  estimation of moving objects for automated rendezvous \& docking,'' in
  \emph{IEEE Int. Conf. on Mechatronics and Automation (ICMA)}, Xian, China,
  August 2010, pp. 305--311.

\bibitem{Zergeroglu-Dawson-2001}
E.~Zergeroglu, D.~M. Dawson, M.~S. de~Querioz, and A.~Behal, ``Vision-based
  nonlinear tracking controllers with uncertain robot-camera parameters,''
  \emph{IEEE/ASME Transactions on Mechatronics}, vol.~6, no.~3, pp. 322--337,
  Sept 2001.

\bibitem{Baeten-DeSchutter-2002}
J.~Baeten and J.~D. Schutter, ``Hybrid vision/force control at corners in
  planar robotic-contour following,'' \emph{IEEE/ASME Transactions on
  Mechatronics}, vol.~7, no.~2, pp. 143--151, June 2002.

\bibitem{Dean-Leon-Parra-Vega-2006}
E.~C. Dean-Leon, V.~Parra-Vega, and A.~Espinosa-Romero, ``Visual servoing for
  constrained planar robots subject to complex friction,'' \emph{IEEE/ASME
  Transactions on Mechatronics}, vol.~11, no.~4, pp. 389--400, Aug 2006.

\bibitem{Aghili-Buehler-Hollerbach-1997a}
F.~Aghili, M.~Buehler, and J.~M. Hollerbach, ``Dynamics and control of
  direct-drive robots with positive joint torque feedback,'' in \emph{{IEEE}
  Int. Conf. Robotics and Automation}, vol.~11, 1997, pp. 1156--1161.

\bibitem{Cheah-Hou-Zhao-2010}
C.~C. Cheah, S.~P. Hou, Y.~Zhao, and J.~E. Slotine, ``Adaptive vision and force
  tracking control for robots with constraint uncertainty,'' \emph{IEEE/ASME
  Transactions on Mechatronics}, vol.~15, no.~3, pp. 389--399, June 2010.

\bibitem{Cai-Dean-2013}
C.~Cai, E.~Dean-Leon, D.~Mendoza, N.~Somani, and A.~Knoll, ``Uncalibrated 3d
  stereo image-based dynamic visual servoing for robot manipulators,'' in
  \emph{2013 IEEE/RSJ International Conference on Intelligent Robots and
  Systems}, Nov 2013, pp. 63--70.

\bibitem{Xie-Low-He-2017}
H.~Xie, K.~H. Low, and Z.~He, ``Adaptive visual servoing of unmanned aerial
  vehicles in gps-denied environments,'' \emph{IEEE/ASME Transactions on
  Mechatronics}, vol.~22, no.~6, pp. 2554--2563, Dec 2017.

\bibitem{Wang-Lang-Silva-2010}
Y.~Wang, H.~Lang, and C.~W. de~Silva, ``A hybrid visual servo controller for
  robust grasping by wheeled mobile robots,'' \emph{IEEE/ASME Transactions on
  Mechatronics}, vol.~15, no.~5, pp. 757--769, Oct 2010.

\bibitem{Mcfadyen-Corke-2014}
A.~Mcfadyen, P.~Corke, and L.~Mejias, ``Visual predictive control of spiral
  motion,'' \emph{IEEE Transactions on Robotics}, vol.~30, no.~6, pp.
  1441--1454, Dec 2014.

\bibitem{Keshmiri-Xie-2017}
M.~Keshmiri and W.~Xie, ``Image-based visual servoing using an optimized
  trajectory planning technique,'' \emph{IEEE/ASME Transactions on
  Mechatronics}, vol.~22, no.~1, pp. 359--370, Feb 2017.

\bibitem{Chen-Wang-Zhao-2018}
R.~Chen, G.~Wang, J.~Zhao, J.~Xu, and K.~Chen, ``Fringe pattern based
  plane-to-plane visual servoing for robotic spray path planning,''
  \emph{IEEE/ASME Transactions on Mechatronics}, vol.~23, no.~3, pp.
  1083--1091, June 2018.

\bibitem{Inaba-Oda-Hayashi-2003}
N.~Inaba, M.~Oda, and M.~Hayashi, ``Visual servoing of space robot for
  autonomous satellite capture,'' \emph{Trans Jpn Soc Aeronaut Space Sci},
  vol.~46, no. 153, pp. 173--9, 2003.

\bibitem{Linchter-Dubowsky-2004}
M.~D. Lichter and S.~Dubowsky, ``State, shape, and parameter estimation of
  space object from range images,'' in \emph{{IEEE} Int. Conf. On Robotics \&
  Automation}, New Orleans, Apr. 2004, pp. 2974--2979.

\bibitem{Aghili-Parsa-2009}
F.~Aghili and K.~Parsa, ``Motion and parameter estimation of space objects
  using laser-vision data,'' \emph{{AIAA} Journal of Guidance, Control, and
  Dynamics}, vol.~32, no.~2, pp. 538--550, March 2009.

\bibitem{Luo-Sakawa-1990}
Z.~Luo and Y.~Sakawa, ``Control of a space manipulator for capturing a tumbling
  object,'' in \emph{Decision and Control, 1990., Proceedings of the 29th IEEE
  Conference on}, Dec. 1990, pp. 103--108 vol.1.

\bibitem{Stengel-1993}
R.~F. Stengel, \emph{Optimal Control and Estimation}.\hskip 1em plus 0.5em
  minus 0.4em\relax New York: Dover Publication, Inc, 1993.

\bibitem{Aghili-2008c}
F.~Aghili, ``Optimal control for robotic capturing and passivation of a
  tumbling satellite with unknown dyanmcis,'' in \emph{{AIAA} Guidance,
  Navigation and Control Conference}, Honolulu, Hawaii, August 2008.

\bibitem{Aghili-2016c}
F.~Aghili and C.~Y. Su, ``Robust relative navigation by integration of icp and
  adaptive kalman filter using laser scanner and imu,'' \emph{IEEE/ASME
  Transactions on Mechatronics}, vol.~21, no.~4, pp. 2015--2026, Aug 2016.

\bibitem{Simon-Herbert-Kanade-1994}
D.~A. Simon, M.~Herbert, and T.~Kanade, ``Real-time 3-d estimation using a
  high-speed range sensor,'' in \emph{{IEEE} Int. Conference on Robotics \&
  Automation}, San Diego, CA, May 1994, pp. 2235--2241.

\bibitem{Besl-Mckay-1992}
P.~J. Besl and N.~D. McKay, ``A method for registration of {3-D} shapes,''
  \emph{{IEEE} Trans. on Pattern Analysis \& Machine Intelligence}, vol.~14,
  no.~2, pp. 239--256, 1992.

\bibitem{Gupta-Hauser-2007}
N.~Gupta and R.~Hauser, ``Kalman filtering with equality and inequality state
  constraints.'' Oxford University Computing Laboratory, University of
  Oxford,http://arxiv.org/abs/0709.2791, Tech. Rep., 2007, technical Report
  07/18, Numerical Analysis Group.

\bibitem{Teixeira-Chandrasekar-2008}
B.~O.~S. Teixeira, J.~Chandrasekar, H.~J. Palanthandalam-Madapusi, L.~A.~B.
  Torres, L.~A. Aguirre, and D.~S. Bernstein, ``Gain-constrained kalman
  filtering for linear and nonlinear systems,'' \emph{IEEE Transactions on
  Signal Processing}, vol.~56, no.~9, pp. 4113--4123, Sept 2008.

\bibitem{Mehra-1970}
R.~Mehra, ``On the identification of variances and adaptive kalman filtering,''
  \emph{Automatic Control, IEEE Transactions on}, vol.~15, no.~2, pp. 175--184,
  Apr 1970.

\bibitem{Wang-2000}
J.~Wang, ``Stochastic modeling for real-time kinematic gps/glonass position,''
  \emph{Journal of Navigation}, vol.~46, no.~4, pp. 297--305, 2000.

\bibitem{Gao-Wei-Zhong-Subic-2015}
S.~Gao, W.~Wei, Y.~Zhong, and A.~Subic, ``Sage windowing and random weighting
  adaptive filtering method for kinematic model error,'' \emph{Aerospace and
  Electronic Systems, IEEE Transactions on}, vol.~51, no.~2, pp. 1488--1500,
  April 2015.

\bibitem{Horn-1987}
B.~K.~P. Horn, ``Closed-form solution of absolute orientation using unit
  quaternions,'' \emph{J. Opt. Soc. Amer.}, vol.~4, no.~4, pp. 629--642, Apr.
  1987.

\bibitem{Anderson-Moore-1990}
B.~D.~O. Anderson and J.~B. Moore, \emph{Optimal Control}.\hskip 1em plus 0.5em
  minus 0.4em\relax Englewood Cliffs, NJ: Prince Hall, 1990.

\bibitem{Boyarko-Yakimenko-2011}
G.~Boyarko, O.~Yakimenko, and M.~Romano, ``Optimal rendezvous trajectories of a
  controled spacecraft and a tumbling object,'' \emph{{AIAA} Jounal of
  Guidance, Control,and Dynamics}, vol.~34, no.~4, pp. 1239--1252, July-August
  2011.

\bibitem{Aghili-Nmavar-2006}
F.~Aghili and M.~Namvar, ``Adaptive control of manipulators using uncalibrated
  joint-torque sensing,'' \emph{IEEE Trans. on Robotics}, vol.~22, no.~4, pp.
  854--860, Aug. 2006.

\end{thebibliography}
%-------------------------------------------------------------

\end{document}